\documentclass[runningheads]{llncs}
\usepackage[misc]{ifsym}
\usepackage[T1]{fontenc}
\usepackage{graphicx}
\usepackage{amsmath,amsfonts}
\usepackage{booktabs}
\usepackage[english]{babel}
\usepackage[table,xcdraw]{xcolor}
\usepackage{appendix}
\usepackage{float}
\usepackage[normalem]{ulem}
\useunder{\uline}{\ul}{}
\usepackage[colorlinks]{hyperref}

\newcommand\blfootnote[1]{%
  \begingroup
  \renewcommand\thefootnote{}\footnote{#1}%
  \addtocounter{footnote}{-1}%
  \endgroup
}

\begin{document}
\definecolor{darkpastelgreen}{rgb}{0.01, 0.75, 0.24}
\title{ReMix: A General and Efficient Framework for Multiple Instance Learning based Whole Slide Image Classification}
\titlerunning{ReMix: A General \& Efficient Framework for WSI Classification}

\author{
Jiawei Yang\inst{1,2,*,\dagger} \and
Hanbo Chen\inst{1,*} \and
Yu Zhao\inst{1} \and
Fan Yang\inst{1} \and
Yao Zhang\inst{3,4} \and
Lei He\inst{2} \and
Jianhua Yao\inst{1}\Letter}
% index{Yang, Jiawei}
% index {Chen, Hanbo}
% index{Zhao, Yu}
% index{Yang, Fan}
% index{Zhang, Yao}
% index{He, Lei}
% index{Yao, Jianhua}

\authorrunning{J. Yang et al.}

\institute{Tencent AI Lab \and University of California, Los Angeles, USA \and Institute of Computing Technology, Chinese Academy of Sciences, Beijing, China \and University of Chinese Academy of Sciences, Beijing, China\\
\email{jiawei118@ucla.edu}}

\maketitle              % typeset the header of the contribution
\blfootnote{*: equally contribute to this work.}
\blfootnote{$\dagger$: work done during an intern at Tencent AI Lab.}
\begin{abstract}
Whole slide image (WSI) classification often relies on deep weakly supervised multiple instance learning (MIL) methods to handle gigapixel resolution images and slide-level labels. Yet the decent performance of deep learning comes from harnessing massive datasets and diverse samples, urging the need for efficient training pipelines for scaling to large datasets and data augmentation techniques for diversifying samples. However, current MIL-based WSI classification pipelines are memory-expensive and computation-inefficient since they usually assemble tens of thousands of patches as bags for computation. On the other hand, despite their popularity in other tasks, data augmentations are unexplored for WSI MIL frameworks. To address them, we propose \texttt{ReMix}, a general and efficient framework for MIL based WSI classification. It comprises two steps: reduce and mix. First, it reduces the number of instances in WSI bags by substituting instances with instance prototypes, \textit{i.e.}, patch cluster centroids. Then, we propose a ``Mix-the-bag'' augmentation that contains four online, stochastic and flexible latent space augmentations. It brings diverse and reliable class-identity-preserving semantic changes in the latent space while enforcing semantic-perturbation invariance. We evaluate \texttt{ReMix} on two public datasets with two state-of-the-art MIL methods. In our experiments, consistent improvements in precision, accuracy, and recall have been achieved but with orders of magnitude reduced training time and memory consumption, demonstrating \texttt{ReMix}'s effectiveness and efficiency. Code is available at \href{https://github.com/TencentAILabHealthcare/ReMix}{https://github.com/TencentAILabHealthcare/ReMix}. 

\end{abstract}

\section{Introduction}
Whole slide images (WSIs) are digital scans of pathology tissue slides that provide critical information for disease diagnosis \cite{srinidhi2020deep}. Recently, many computer-aided WSI diagnostic systems have been developed upon deep learning (DL) methods \cite{hou2016patch,sirinukunwattana2017gland,hashimoto2020multi}. However, some challenges in WSI classification still exist.

First of all, WSIs are huge and challenging to be processed at once by DL models. Given micron-size pixels and centimeter-sized slides, a WSI is usually of gigapixel size. Thus, it needs to be divided into thousands or tens of thousands of small ``patches'' for computational analysis. Current successful paradigms formulate WSI classification as a weakly supervised multiple instance learning (MIL) problem \cite{hou2016patch,hashimoto2020multi,ilse2018attention}. Under this setting, each WSI is regarded as a \textit{bag} that contains many \textit{instances} of patches. In practice, each bag is loaded into memory and fed into deep MIL models separately due to poor parallelization and large size. However, the number of instances in each bag can vary strikingly, leading to unstable input/output (I/O) stream. Besides, the different numbers of instances also make parallelizing MIL models hard as they cannot be directly composed into a batch. One naive solution is to select small random subsets of patches from each WSI \cite{naik2020deep}. Alternatively, Yao~\textit{et al.} \cite{yao2020whole} propose to cluster over the extracted features based on an ImageNet-pre-trained encoder to define phenotype groups and sample patches from those groups each time. 

Second, DL models are usually data-hungry --- they tend to perform better with diversified labeled data, which are expensive to collect. Data augmentation is a simple yet effective approach to improving data diversity \cite{shorten2019survey}. Over the years, many augmentation methods have been proposed for different tasks. However, MIL problems' augmentations are unexplored, especially for WSI classification. Using image processing functions such as cropping, flipping, or color shifting for all instances is extremely inefficient. Therefore, developing efficient and effective augmentation methods for WSI classification is of significant interest.  

To address the issues mentioned above, we propose \texttt{ReMix}, a general and efficient framework for MIL-based WSI classification. It consists of two steps: (1) Reduce and (2) Mix. Given the nature of WSIs that many tissues and contexts are similar, repetitive, and sometimes redundant, we hypothesize that using the clustered instance prototypes, instead of all available patches, to represent a WSI can still yield comparably well or even better performance. By doing so, \texttt{ReMix} \textit{reduces} the number of instances inside a bag by orders of magnitude (\textit{e.g.}, from $10^4$ to $10^0\sim 10^1$). After that, we propose a novel augmentation method called ``Mix-the-bag.'' It contains several online, stochastic and flexible latent space augmentations that are directly applied on bags as well as on instance prototypes. It \textit{mixes} different bags by appending, replacing, interpolating instance prototypes, or transferring semantic variations between different bags. This process is applied in the latent space via numerical addition, which is highly efficient. Our key contributions are:
\begin{itemize}
    \item Propose a general, simple yet effective method to improve the training efficiency of the MIL framework for WSI classification.
    \item Propose a novel and efficient latent augmentation method for MIL-based WSI classification, which is much underexplored in existing works.
    \item Improve previous state-of-the-art MIL methods on two public datasets by considerable margins but with orders of magnitude reduced budgets.
\end{itemize}

\section{Method}

\subsection{Preliminary: MIL Formulation}
In MIL, a dataset that has $N$ bags is typically formulated as $\mathcal{D}=\{(B_i, y_i)\}_{i=1}^{N}$, where $B_i=\{x_k\}_{k=1}^{N_i}$ denotes the $i$-th bag that has $N_i$ instances and $y_i$ is the bag label. For binary classification, a bag is defined as positive if it contains at least one positive instance and negative otherwise. A general \textit{spatial-agnostic} MIL classification process that does not rely on the spatial relationship between instances to make decisions can be expressed as $\hat{y}_i=g\left(f(x_1), ..., f(x_{N_i})\right)$, where $f(\cdot)$ is a patch instance encoder, and $g(\cdot)$ is an instances aggregator that aggregates information and makes final predictions. 

\subsection{ReMix}
%##################################################################################################
\begin{figure}[t!]
\centering
\includegraphics[width=0.9\textwidth]{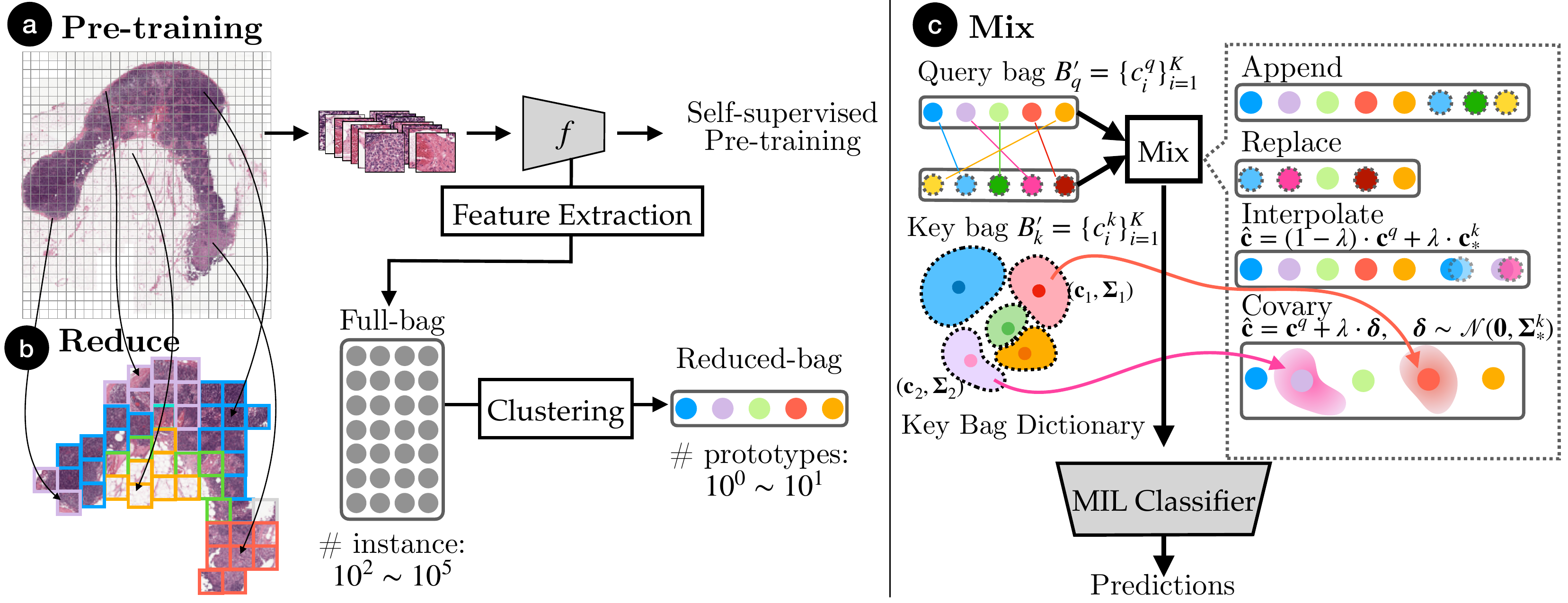}
\caption{\textbf{\texttt{ReMix}'s overview.} (a) Patch encoder pre-training. (b) Reduce the number of instances by substituting them with prototypes (right); several patches can abstract a large-size whole slide image (left). (c) Mix the bags by appending, replacing, interpolating prototypes, or transferring intra-cluster covariance from other WSIs.} 
\label{fig:overview}
\end{figure}
%##################################################################################################

\subsubsection{Overview.} 
Instead of improving the performance of a specific dataset or model, our \texttt{ReMix} is a general framework that can accommodate most spatial-agnostic MIL methods. Figure \ref{fig:overview} shows its overview. It consists of two efficient and effective steps, \textit{i.e.}, reduce and mix, which we elaborate on below. 

\subsubsection{Pre-training.} 
Due to the weakly supervision nature of MIL problems, patch encoder training faces the issue of lacking appropriate patch-level labels. Existing end-to-end training methods are usually expensive \cite{campanella2019clinical,li2021dual}, inefficient, and sometimes infeasible. Using a pre-trained encoder has become a common practice \cite{li2021dual,yao2020whole}, such as an ImageNet pre-trained encoder, a pseudo fully-supervised pre-trained encoder, or a self-supervised pre-trained encoder. The pseudo fully-supervised pre-training \cite{hou2016patch,campanella2019clinical,hashimoto2020multi,chen2019rectified} assigns the bag labels to all instances inside the bags and conducts patch classification pre-training. The self-supervised learning methods such as SimCLR \cite{chen2020simple} and MoCo \cite{he2020momentum} obtain good pre-trained models by maximizing the similarity between two different augmented views from the same patch while minimizing it between views from different patches. Recent studies have witnessed the superiority of self-supervised pre-training on large-scale and imbalanced WSI patches over others \cite{li2021dual,ciga2022self,yang2022towards}. We follow \cite{li2021dual} to adopt a state-of-the-art self-supervised learning method --- SimCLR \cite{chen2020simple} for patch encoder pre-training. Note that the choice of patch-encoder is orthogonal to the \texttt{ReMix} framework and downstream MIL models' training. We here briefly discuss available pre-training methods only for completeness. Thus their training budgets are not considered in this work.

\subsubsection{Reduce.} 
\texttt{ReMix} reduces the number of instances in each bag via clustering. Conventionally, all patches extracted from a WSI are assembled as a bag for downstream MIL classification \cite{li2021dual,hou2016patch}. However, the number of instances inside a bag can be huge (\textit{e.g.}, $N_i=50,000$), leading to heavy I/O cost and high memory consumption during training. Given the nature of histological scans in mind that a large portion of tissues is usually similar, repetitive and, sometimes redundant, we propose to substitute instances with instance prototypes. Specifically, for each bag, we perform K-Means on the patches' representations to obtain $K$ clusters and use their prototypes (centroids) to represent the bag, \textit{i.e.}, $B'_i=\{\mathbf{c}_k\}_{k=1}^K$, where $\mathbf{c}_k$ corresponds to the $k$-th prototype. Below, we denote the reduced-bags as $B'$ and the full-bags as $B$. The WSI thumbnails in Figure \ref{fig:overview}-(b) illustrate how several patches (reduced-bag) can provide sufficient information of the entire WSI (full-bag) for certain downstream tasks, \textit{e.g.}, whole-slide classification. The reduced-bag can be seen as a denoised abstraction of full-bag.

To fully exploit WSI information, inspired by \cite{yang2022towards}, we construct a \textit{bag dictionary} as $\Phi_i=\{(\mathbf{c}_k, \mathbf{\Sigma}_k)\}_{k=1}^{K}$ for each bag, where $\mathbf{\Sigma}_k$ corresponds to the intra-cluster covariance matrix of the $k$-th cluster. A bag dictionary captures how its instances distribute at a high level by modeling a multivariate Gaussian distribution $\mathcal{N}(\mathbf{c}_k, \mathbf{\Sigma}_k)$. Besides, the covariance can reflect the semantic directions inherent in each cluster, \textit{e.g.}, how features vary in that cluster.

\subsubsection{Mix.} 
\texttt{ReMix} applies latent space augmentation by mixing the bags to increase data diversity. DL models are prone to overfit with limited labeled training samples. Data augmentation can provide additional artificial data \cite{cheung2020modals,ghiasi2020simple,shorten2019survey,sohn2020fixmatch}. However, using image processing functions for bags can be extremely inefficient. Here we propose ``mix-the-bag'' augmentation that is illustrated in Figure \ref{fig:overview}-(c). At a high level, when a bag is fed into a MIL classifier, we randomly sample another bag of the same class and ``mix'' them. Without loss of generality, we define the former bag as a query bag $B'_q=\{\mathbf{c}_i^q\}_{i=1}^K$ and the latter bag as a key bag $B'_k=\{\mathbf{c}_i^k\}_{i=1}^{K}$. Their instances $\mathbf{c}^q$ and $\mathbf{c}^k$ are called query prototypes and key prototypes subsequently. For each query prototype $\mathbf{c}_i^q$, we find its closest key prototype $\mathbf{c}_{i^*}^k$ and then augment the query bag with an applying probability of $p$ by one of the following four augmentations:
\begin{itemize}
	\item \textbf{Append:} append the closest key prototype $\mathbf{c}_{i^*}^k$ to query bag $B_q$.
	\item \textbf{Replace:} replace the query prototype $\mathbf{c}_{i}^q$ with its closest key prototype $\mathbf{c}_{i^*}^k$.
	\item {\textbf{Interpolate:} append an interpolated representation
			\begin{equation}
				\hat{\mathbf{c}}_i=(1-\lambda) \cdot \mathbf{c}_{i}^q + \lambda \cdot \mathbf{c}_{i^*}^k
			\end{equation}
			to the query bag $B_q$, where $\lambda$ is a strength hyper-parameter.}
	\item {\textbf{Covary:} generate a new representation from the key covariance matrix by
			\begin{equation}
				\hat{\mathbf{c}}_i = \mathbf{c}_{i}^q + \lambda \cdot \boldsymbol{\delta}, \quad \boldsymbol{\delta} \sim \mathcal{N}(\mathbf{0}, \mathbf{\Sigma}_{i^*}^k)
			\end{equation}
			and append it to the bag $B_q$, where $\lambda$ is a strength hyper-parameter and $\mathbf{\Sigma}_{i^*}^k$ is the covariance matrix corresponding to the closest key prototype $\mathbf{c}_{i^*}^k$.}
\end{itemize}
It is vital to sample another bag from the \textit{same class} and mix the query prototype with the \textit{most similar} key prototype. It helps preserve critical class-related information and reduces the risk of losing original class identity. In addition to four individual augmentations, we propose to combine them sequentially as a ``\textbf{joint}'' augmentation.

\subsection{Intuitions on \texttt{ReMix}'s Effectiveness}

\subsubsection{Implicit data re-balance behavior.} 
Tissue imbalance is a common problem with WSIs. The majority of similar patches almost convey the same information about the WSI but could dominate in numbers over other distinct minority patches. Using the representative prototypes to assemble WSI bags can be seen as an implicit data re-balance mechanism that bridges the gap between the numbers of the majority and the minority. It alleviates the tissue imbalance problem to some extent. Besides, using the mean embedding of a group of similar patches could obtain a more accurate and less noisy tissue representation.

\subsubsection{Efficient semantic consistency regularization.} 
Consistency regularization underlies many successful works, such as semi-supervised learning \cite{berthelot2019mixmatch,berthelot2019remixmatch,sohn2020fixmatch}. Usually, consistency regularization enforces models' predictions to be invariant under different data augmentations \cite{sohn2020fixmatch}. Instead of augmenting instances using image processing functions in the input RGB space, \texttt{ReMix} augments the bags by bringing diverse semantic changes in the latent space. Guided by bag labels and prototypes similarity, such changes are class-identity-preserving. The bag instance combination is no longer static and unaltered but diverse and dynamic, \textit{i.e.}, different new bags can be fed into the MIL classifier every time. Our augmentations can be seen as efficient semantic consistency regularization methods.

\subsubsection{Why clustering and additive latent augmentation works.} 
When learned properly, the deep representation space is shown to be highly linearized \cite{bengio2013better,upchurch2017deep}. Consequently, the distance metrics could demonstrate the similarity between patches, making clustering meaningful. Moreover, in such a space, linear transformation, \textit{e.g.}, interpolating features or adding semantic translation vector $\boldsymbol{\delta}$, is likely to provide plausible representations \cite{cheung2020modals}. The mixed bag representations can serve as hard examples that help models generalize better \cite{kuchnik2018efficient,zhang2019adversarial,wu2020generalization}.

\section{Experiments}

\subsection{Datasets and Metrics}

\subsubsection{UniToPatho.} 
UniToPatho \cite{unitopatho} is a public dataset comprising 9536 hematoxylin and eosin (H\&E) stained patches extracted from 292 WSIs. The slides are scanned at 20$\times$ magnification (0.4415 $\mu$m/px). There are six classes in this dataset. We use the official split of 204/88 slides for training/testing, respectively. For patch processing, we crop the provided images into $224 \times 224$ sized patches without overlapping. Under this setting, the average number of instances per bag is about 1.6k, with the largest bag having more than 20k instances.

\subsubsection{Camelyon16.} 
Camelyon16 \cite{bejnordi2017diagnostic} is a publicly available dataset consisting of 400 H\&E stained slides from breast cancer screening. It contains two classes, \textit{i.e.}, normal and tumor. We directly use the pre-computed features provided by DSMIL \cite{li2021dual} without further processing. There are 271/129 slides in the training/testing set. The average number of instances per bag is around 8k, with the largest bag having  more than 50k instances.

\subsubsection{Metrics.} 
We report class-wise averaged precision, recall, and accuracy. To avoid randomness, we run all experiments ten times and report the averaged metrics. 

\subsection{Implementation Details}

\subsubsection{Patch encoder.}
For the Camelyon16 dataset \cite{bejnordi2017diagnostic}, we use the pre-computed features provided by \cite{li2021dual}. They pre-train a SimCLR patch encoder directly on the Camelyon16 dataset and use it for Camelyon16 patch encoding. To align with that setting, we also pre-train a ResNet-18 encoder \cite{he2016deep} with SimCLR \cite{chen2020simple} on UniToPatho \cite{unitopatho} dataset. More pre-training details are in Supplementary.
	
\subsubsection{MIL models.} 
To demonstrate that \texttt{ReMix} can be MIL model-agnostic, we use two previous state-of-the-art deep MIL models, namely ABMIL \cite{ilse2018attention} and DSMIL \cite{li2021dual}, for our experiments. ABMIL and DSMIL are both attention-based MIL methods that compute the attention-weighted sum of instances features as the bag representation. They differ in the way of attention computing. ABMIL \cite{ilse2018attention} predicts the attention scores of each patch using a multi-layer perceptron (MLP) without explicit patch relation modeling. DSMIL \cite{li2021dual} is a dual-stream method that comprises an instance branch and a bag branch. The instance branch identifies the highest scored instance. In addition, the bag branch measures the similarity between other patches and the highest scored instance and thereafter utilizes the similarity scores to compute attention. 
    
We use DSMIL's codebase for implementation and training. Unless other specified, all MIL models are optimized for 50 epochs by Adam optimizer \cite{kingma2014adam} with an initial learning rate of 2e-4 and a cosine annealing learning rate schedule \cite{loshchilov2016sgdr}. The mini-batch size is 1 (bag) for a fair comparison, despite that \texttt{ReMix} can easily scale it up since the reduced bags have the same number of instances and thus can be composed into a batch for parallel computing. 

\subsubsection{\texttt{ReMix} hyper-parameters.} 
There are three hyper-parameters in \texttt{ReMix}, \textit{i.e.}, number of prototypes $K$, augmentation probability $p$, and strength $\lambda$. To study the effects of different hyper-parameters, we first sweep $K$ in $\{1, 2, 4, 8, 16, 32\}$ to find the optimal $K$ for each method and dataset. For simplicity and bag diversity, we set $p=0.5$ and uniformly sample $\lambda$ from $(0,1)$ in all individual augmentations. For ``joint'' augmentation, we set $p=0.1$ since it comprises four augmentations and too strong augmentation is known to be harmful. Both MIL methods share the optimal $K$ values: $K=1$ for the UniToPatho dataset and $K=8$ for the Camelyon16 dataset. Due to limited space, we postpone the empirical studies for each hyper-parameter to Supplementary, \textit{e.g.}, studying the robustness to the choice of augmentation probability $p$ and the choice of the number of prototypes $K$.

\subsection{Comparisons}

%##################################################################################################
\begin{table}[t]
\begin{center}
\caption{\textbf{Main results.} The ``Average'' columns report the mean of precision, recall and accuracy. Bold, underlined and italicized numbers are the first, second and third best entries. All results are averaged over 10 independent runs. Numbers are shown in percentage (\%). ``no aug.'' means no augmentation.}
\label{tab:main_results}
\resizebox{0.9\textwidth}{!}{%
\begin{tabular}{@{}l|cccc|cccc@{}}
\toprule
      & \multicolumn{4}{c|}{UniToPatho Dataset \cite{unitopatho}}         & \multicolumn{4}{c}{Camelyon16 Dataset \cite{bejnordi2017diagnostic}}          \\ \midrule
    Methods\textbackslash Metrics  & Precision & Recall & Accuracy & Average & Precision & Recall & Accuracy & Average \\ \midrule
ABMIL \cite{ilse2018attention} (baseline)   & 56.18     & 58.50  & 60.11    & 58.26 
                            & 92.47         & 92.79                 & 93.02          & 92.76   \\
ReMix-ABMIL (no aug.)       & 69.93         & 72.85                 & 68.75          & 70.51 
                            & 93.97         & 93.15                 & 93.95          & 93.69   \\
ReMix-ABMIL (append)        & \textit{71.81}& 74.54                 & 69.09          & 71.81 
                            & 94.59         & 93.38                 & \textit{94.34}          & 94.10 \\
ReMix-ABMIL (replace)       & 70.16         & 74.34                 & 68.75          & 71.08 
                            & 94.60         & {\ul 93.52}           & {\ul 94.42}    & \textit{94.18}   \\
ReMix-ABMIL (interpolate)   & 71.55         & \textit{75.54 }       & \textit{70.23}          & \textit{72.44}   
                            & \textit{94.65}   & \textit{93.49}     & {\ul 94.42}    & {\ul 94.19}  \\
ReMix-ABMIL (covary)        & \textbf{72.32} & \textbf{76.71}       & \textbf{71.02} & \textbf{73.35} 
                            & \textbf{94.75} & \textbf{93.55}       & \textbf{94.49} & \textbf{94.26}   \\ 
ReMix-ABMIL (joint)        & {\ul 72.13}    & {\ul 76.00}           & {\ul 70.91} & {\ul 73.01} 
                            & {\ul 94.69} & {93.45}       & {\ul 94.42} & \textit{94.18}   \\ 
\midrule
Best Improvement $\mathrm{\Delta}$ 
            & {\color{darkpastelgreen} \textbf{+16.14}} 
            & {\color{darkpastelgreen} \textbf{+18.21}} 
            & {\color{darkpastelgreen} \textbf{+10.91}} 
            & {\color{darkpastelgreen} \textbf{+15.09}} 
            & {\color{darkpastelgreen} \textbf{+2.28}} 
            & {\color{darkpastelgreen} \textbf{+0.76}}  
            & {\color{darkpastelgreen} \textbf{+1.47}} 
            & {\color{darkpastelgreen} \textbf{+1.50}} \\
	\midrule \midrule
DSMIL \cite{li2021dual} (baseline)        & 72.92     & 79.41  & 76.36    & 76.23   &
						  94.37     & 93.39  & 94.11$^\dagger$    & 93.96   \\
ReMix-DSMIL (no aug.)       & 76.14     & 79.26  & 77.95    & 77.78   
                            & 95.68     & 93.44  & 94.80    & 94.64   \\
ReMix-DSMIL (append)        & {\ul 77.91} & {\ul 80.56} & \textbf{81.02} & {\ul 79.83} 
                            & {\ul 96.39}    & \textbf{94.10} & \textbf{95.43} & \textbf{95.31}   \\
ReMix-DSMIL (replace)       & 76.60          & 79.30          & 78.64          & 78.18          
                            & 95.33          & 93.44          & 94.65          & 94.47   \\
ReMix-DSMIL (interpolate)   & 76.99          & 80.26          & 80.00          & 79.08
                            & {\ul 96.39}          & 93.96          & 95.35          & 95.23   \\
ReMix-DSMIL (covary)        & \textit{77.72}    & \textit{80.52}    & \textit{80.46}    & \textit{79.57}    
                            & \textbf{96.51} & \textit{93.88}    & {\ul 95.35}    & {\ul 95.25}   \\ 
ReMix-DSMIL (joint)         & \textbf{78.20}    & \textbf{80.94}    & {\ul80.68}    & \textbf{79.94}    
                            & \textit{96.18} & {\ul 93.97}    & \textit{95.27}    & \textit{95.14}   \\ \midrule
Best Improvement $\mathrm{\Delta}$ 
            & {\color{darkpastelgreen} \textbf{+5.28}} 
            & {\color{darkpastelgreen} \textbf{+1.53}} 
            & {\color{darkpastelgreen} \textbf{+4.66}} 
            & {\color{darkpastelgreen} \textbf{+3.71}} 
            & {\color{darkpastelgreen} \textbf{+2.14}}  
            & {\color{darkpastelgreen} \textbf{+0.71}}  
            & {\color{darkpastelgreen} \textbf{+1.32}} 
            & {\color{darkpastelgreen} \textbf{+1.35}} \\  \bottomrule
\end{tabular}%
}
\end{center}
\scriptsize{$^\dagger$The reported accuracy for Camelyon16 in DSMIL \cite{li2021dual} is 89.92\%. We reproduce a better baseline.}
\end{table}
%##################################################################################################

\subsubsection{Main results.}
Table \ref{tab:main_results} shows the main results. Even without ``mix-the-bag'' augmentations (no aug.), \texttt{ReMix} can improve previous state-of-the-arts by only the ``reduce'' step in both datasets, \textit{e.g.}, +13.75\% and +3.22\% precision for ABMIL and DSMIL respectively in UniToPatho and +1.50\%, +1.31\% precision for them in Camelyon16. Overall, ABMIL benefits more from \texttt{ReMix} than DSMIL. DSMIL computes \textit{self-attention} that explicitly considers the similarity between different instances inside a bag, while ABMIL directly predicts attention scores using an MLP for all instances without such explicit inter-instance relation computing. For this reason, we conjure that ABMIL's attentions are more likely to overfit on redundant instances than DSMIL's, and thus it benefits more from the denoised reduced-bags. Using the representative prototypes can ease the recognition process and alleviate the overfitting problem. These results suggest that \texttt{ReMix} can reduce data noise in the bag representations to some extent and improve performance. 

Applying ``mix-the-bag'' augmentations can further improve the performance of reduced-bags (no aug.) by a considerable margin, \textit{e.g.}, +2.27\% and +3.07\% accuracy for ReMix-ABMIL and ReMix-DSMIL respectively, in UniToPatho. In general, append-augmentation fits more to DSMIL, while ABMIL favors covary-augmentation. Especially, covary-augmentation achieves the top-2 performance across datasets, confirming our motivation that transferring others' covariance in the latent space could provide reliable and diversified variations for semantic augmentation. 
Using full-bags (baseline) can be seen as a special case of augmenting the prototypes with their own covariances. However, such bags are static and unaltered as discussed in \S 2.3. In contrast, with \texttt{ReMix}, the reduced and augmented bags can be more diverse and dynamic. Such augmentations are important for low-data applications, \textit{e.g.}, WSI classification. Using ``joint'' augmentation can further improve some of the results but not all. It is anticipated and in line with experience from other tasks and fields that too strong augmentation is not beneficial for training (\textit{e.g.}, Table 2 from \cite{appalaraju2020towards}).

Overall, solid gains observed in Table \ref{tab:main_results} have confirmed the effectiveness of the proposed \texttt{ReMix} framework. We next demonstrate its efficiency.

\subsubsection{Training budgets.} 
We compare the training budgets, \textit{i.e.}, average training time per epoch and the peak memory consumed during training, in Table \ref{tab:comp_budget}. Although all models are trained with a batch size of 1 for a fair comparison, the \texttt{ReMix} framework significantly outperforms other entries in all training budgets. It costs nearly $20 \times$ less training time but obtains better results for both MIL methods in UniToPatho (\textit{e.g.}, +10.91\% accuracy). Moreover, it takes about at least $200\times$ shorter training time to achieve better results than the original ones in the Camelyon16 dataset, whose average bag size is about $5\times$ as big as UniToPatho's. It can be expected that the training efficiency gains owned by \texttt{ReMix} would enlarge as the bag size increases. With more data collected in the real world, we argue that the training efficiency should be as important as the classification performance when scaling up to large datasets. Therefore, we emphasize the superiority of \texttt{ReMix} in being an extremely efficient framework.

\subsubsection{More studies.} We provide more empirical studies, \textit{i.e.}, the effect of the number of prototypes $K$, the effect of augmentation probability $p$, and the effect of training MIL classifier longer, in Supplementary to better understand \texttt{ReMix}.

%##################################################################################################
\begin{table}
\begin{center}
\caption{\textbf{Comparison of training budgets.} 
    Numbers are estimated from 50-epoch training on the same machine with an 8GB Tesla T4-8C virtual GPU. 
    ``Original / ReMix'' rows show the multiplier between the original's and ReMix improved budgets.}
\label{tab:comp_budget}
\resizebox{0.7\textwidth}{!}{%
\begin{tabular}{@{}l|cc|cc@{}}
\toprule
& \multicolumn{2}{c|}{Average Seconds / Epoch} & \multicolumn{2}{c}{Memory   Peak} \\ \midrule
Methods \textbackslash Datasets & UniToPatho  & Camelyon16  & UniToPatho    & Camelyon16        \\ \midrule
ABMIL$^\dagger$ (full-bag) & 18.41$''$    &  235.72$''$    & 55.63 MB         &  332.12 MB        \\
ReMix-ABMIL (no aug.)          & 0.84$''$    &  1.10$''$      & 6.45 MB      &  8.76 MB         \\ \midrule
Original / ReMix           & 21.93$\times$  &  214.29$\times$     & 8.61$\times$  &  37.91 $\times$  \\ \midrule \midrule
DSMIL$^\dagger$ (full-bag) & 19.20$''$     &  255.14$''$   & 66.58 MB      &   364.72 MB    \\
ReMix-DSMIL (no aug.)          & 0.85$''$     &  1.12$''$    & 6.46 MB       &   8.76 MB  \\ \midrule
Original / ReMix           & 22.57$\times$ &  227.80$\times$ & 10.31$\times$    &  41.63 $\times$  \\ \bottomrule
\end{tabular}%
}
\end{center}
\scriptsize{$^\dagger$We have already improved the original DSMIL/ABMIL training by at least 2$\times$ faster speed and 2.4$\times$ less memory consumption, \textit{e.g.}, the original recipe for Camelyon16 is $552.7''$/epoch and 800.35MB. Note that, we use a distributed cluster storage platform (Ceph), so the data loading time might be longer than that in local disks.}
\end{table}
%##################################################################################################

\section{Conclusion}
This work presents \texttt{ReMix}, a general and efficient framework for MIL-based WSI classification. \texttt{ReMix} reduces the number of instances in WSI bags by substituting instances with instance prototypes, and mixes the bags via latent space augmentation methods to improve data diversity. Our \texttt{ReMix} can considerably improve previous state-of-the-art MIL classification methods, yet with faster training speed and less memory consumption, showing its effectiveness and efficiency.

 \bibliographystyle{splncs04}
 \bibliography{mybibliography}

\newpage
\appendix

\setcounter{figure}{0}
\renewcommand{\thefigure}{A.\arabic{figure}}

\setcounter{table}{0}
\renewcommand{\thetable}{A.\arabic{table}}

\section{Supplementary Materials} 
\label{sec:empirical_study}

\begin{table}[]
\caption{\textbf{SimCLR pre-training settings.}}
\label{tab:pretraining}
\resizebox{\textwidth}{!}{%
\begin{tabular}[c]{@{}ll@{}}
\toprule
\textbf{Configuration}         & \textbf{Detailed setting}                              \\ \midrule
Codebase                    & \begin{tabular}[t]{@{}l@{}}OpenSelfsup\\(Current address: \url{https://github.com/open-mmlab/mmselfsup})\end{tabular}                                              \\ \hline
Model                       & 
SimCLR with ResNet18 and a two-layer non-linear projection head \\\hline
Temperature $\tau$ & 0.1                                                          \\\hline
Normalization               & 
%\begin{tabular}[t]{@{}l@{}}ImageNet statistics, \textit{i.e.}, mean=(0.485, 0.456, 0.406) \\and std=(0.229, 0.224, 0.225).\end{tabular}     
Using ImageNet mean and std.
\\\hline
Augmentation                
& 
% \begin{tabular}[t]{@{}l@{}}
% \texttt{RandomResizedCrop} to 224$\times$224,\\
% \texttt{RandomHorizontalFlip}, \\
% \texttt{ColorJitter} with an applying probability of 0.8 in the ranges of \\brightness 0.8, contrast 0.8, saturation 0.8 and hue 0.2, \\
% \texttt{RandomGrayscale} of an applying probability of 0.2, \\
% \texttt{GaussianBlur} with $\sigma_{min}=0.1$ and $\sigma_{max}=2.0$ with probability of 0.5
% \end{tabular} 
Following the original configs in the codebase
\\\hline
Optimizer                   
& \texttt{LARS} optimizer with an initial $lr$ of 0.6, $wd$ of 1e-6, and momentum of 0.9                                                                  \\\hline
Schedule                    & \texttt{CosineAnnealing} learning rate scheduler with 10-epoch warm-up                                                                      \\\hline
Training iterations         & 200 epochs with a batch size of 512                                      \\ \bottomrule
\end{tabular}
}
\end{table}
	
\begin{figure}
\centering
\includegraphics[width=\textwidth]{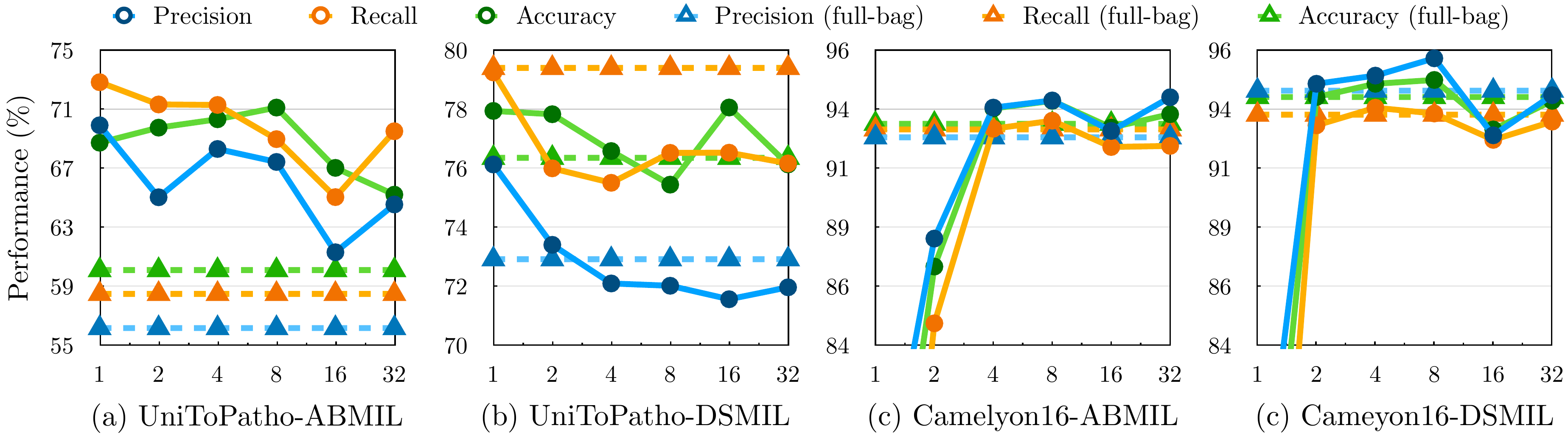}
\caption{\textbf{
Empirical study on the number of prototypes.} 
Horizontal axes denote the number of prototypes in the reduced-bags. Baselines are the results trained on the full-bags. Results are the average of 10 runs.
In UniToPatho, both ABMIL and DSMIL achieve the best results with 1 prototype. 
ABMIL is robust to the choice of $K$ as it consistently outperforms full-bags with the reduced-bag representations. 
In Camelyon16, \texttt{ReMiX} performs similarly well when $K\geq4$, with $K=8$ being the best. 
Tumors' lesion area accounts for only 10\% to 30\% in Camelyon16. 
Thus, more prototypes are needed for bag representation. 
Nevertheless, reduced-bags ($10^0\sim 10^1$ instances/bag) are still significantly cheaper than full-bags ($10^3 \sim 10^4$ instances/bag).
} 
\label{fig:num_protos}
\end{figure}

\begin{table}[h!]
\centering
\caption{\textbf{Empirical study on training epochs.} We show the average of precision, recall and accuracy. All results are averaged over 10 trials with standard deviations ($\pm$). Numbers are shown in \%.
 When trained on full-bags, both MIL methods do not gain much from longer training. 
The performance of ABMIL even drops considerably with 200-epoch training. 
Surprisingly, when trained on reduced-bags, ABMIL starts to benefit from longer training, showing its better potential. 
Overall, all of the tested cases support the superiority of \texttt{ReMix} regardless the number of training epochs.
}
\label{tab:longer}
\resizebox{0.7\textwidth}{!}{%
\begin{tabular}{@{}c|cc|cc@{}}
\toprule
             & \multicolumn{2}{c|}{UniToPatho-ABMIL} & \multicolumn{2}{c}{UniToPatho-DSMIL} \\ \midrule
Epoch        & Full-bag   		& Reduced-bag   	 & Full-bag   	   & Reduced-bag  \\ \midrule
50 (default) & 58.26$\pm$6.64 & 70.51$\pm$2.54   & 76.23$\pm$0.64  & 77.78$\pm$1.09        \\
100          & 58.95$\pm$2.17 & 74.24$\pm$0.62   & 76.39$\pm$0.08  & 78.50$\pm$1.36         \\
200          & 55.60$\pm$3.16 & 76.14$\pm$1.23   & 76.80$\pm$0.31  & 77.92$\pm$0.60        \\ \bottomrule
\end{tabular}%
}
\end{table}

\begin{table}[h!]
\centering
\caption{\textbf{Empirical study on augmentation probabilities.} 
The displayed metrics are the average of precision, recall and accuracy. All results are averaged over 10 runs. Numbers are shown in percentage (\%).
We estimate the expected performance given an augmentation with varying probability ($\mathbb{E}(\mathrm{aug}|p)$) and the expected performance given a fixed probability with varying augmentations ($\mathbb{E}(p|\mathrm{aug})$). 
The results indicate \texttt{ReMix} robustness to the choice of augmentation probability. 
}
\label{tab:prob_exp}
\resizebox{0.96\textwidth}{!}{%
\begin{tabular}{l|cccccc||cccccc}
\hline
 &
  \multicolumn{6}{c|}{UniToPatho-ABMIL} &
  \multicolumn{6}{c}{UniToPatho-DSMIL} \\ \hline\hline
  Augs.\textbackslash Prob. &
  0.1 &
  0.3 &
  0.5 &
  0.7 &
  \multicolumn{1}{c|}{0.9} &
  $\mathbb{E}(\mathrm{aug}|p)$ &
  0.1 &
  0.3 &
  0.5 &
  0.7 &
  \multicolumn{1}{c|}{0.9} &
  $\mathbb{E}(\mathrm{aug}|p)$ \\ \hline
baseline (full-bag) &
  \multicolumn{5}{c|}{58.26} &
  \cellcolor[HTML]{EFEFEF}58.26 &
  \multicolumn{5}{c|}{76.23} &
  \cellcolor[HTML]{EFEFEF}76.23 \\
ReMix (no aug) &
  \multicolumn{5}{c|}{70.51} &
  \cellcolor[HTML]{EFEFEF}70.51 &
  \multicolumn{5}{c|}{77.78} &
  \cellcolor[HTML]{EFEFEF}77.78 \\ \hline
ReMix(append) &
  71.69 &
  71.12 &
  71.81 &
  72.23 &
  \multicolumn{1}{c|}{70.11} &
  \cellcolor[HTML]{EFEFEF}71.39 &
  78.64 &
  79.15 &
  79.83 &
  78.87 &
  \multicolumn{1}{c|}{78.04} &
  \cellcolor[HTML]{EFEFEF}78.91 \\
ReMix(replace) &
  71.25 &
  70.72 &
  71.08 &
  70.94 &
  \multicolumn{1}{c|}{71.19} &
  \cellcolor[HTML]{EFEFEF}71.04 &
  78.13 &
  77.80 &
  78.18 &
  77.68 &
  \multicolumn{1}{c|}{77.86} &
  \cellcolor[HTML]{EFEFEF}77.93 \\
ReMix(interpolate) &
  71.45 &
  72.67 &
  72.44 &
  72.62 &
  \multicolumn{1}{c|}{71.31} &
  \cellcolor[HTML]{EFEFEF}72.10 &
  78.73 &
  79.21 &
  79.08 &
  78.24 &
  \multicolumn{1}{c|}{77.60} &
  \cellcolor[HTML]{EFEFEF}78.57 \\
ReMix(covary) &
  72.15 &
  71.78 &
  73.35 &
  70.73 &
  \multicolumn{1}{c|}{73.00} &
  \cellcolor[HTML]{EFEFEF}72.20 &
  78.86 &
  77.92 &
  79.57 &
  78.51 &
  \multicolumn{1}{c|}{78.61} &
  \cellcolor[HTML]{EFEFEF}78.69 \\ \hline
$\mathbb{E}(p|\mathrm{aug})$ &
  \cellcolor[HTML]{EFEFEF}71.64 &
  \cellcolor[HTML]{EFEFEF}71.57 &
  \cellcolor[HTML]{EFEFEF}72.17 &
  \cellcolor[HTML]{EFEFEF}71.63 &
  \multicolumn{1}{c|}{\cellcolor[HTML]{EFEFEF}71.40} &
  \cellcolor[HTML]{EFEFEF}71.68 &
  \cellcolor[HTML]{EFEFEF}78.59 &
  \cellcolor[HTML]{EFEFEF}78.52 &
  \cellcolor[HTML]{EFEFEF}79.17 &
  \cellcolor[HTML]{EFEFEF}78.33 &
  \multicolumn{1}{c|}{\cellcolor[HTML]{EFEFEF}78.03} &
  \cellcolor[HTML]{EFEFEF}78.53 \\
  \hline
  \hline
 &
  \multicolumn{6}{c|}{Camelyon16-ABMIL} &
  \multicolumn{6}{c}{Camelyon16-DSMIL} \\ \hline\hline
baseline (full-bag) &
  \multicolumn{5}{c|}{92.76} &
  \cellcolor[HTML]{EFEFEF}92.76 &
  \multicolumn{5}{c|}{93.96} &
  \cellcolor[HTML]{EFEFEF}93.96 \\
ReMix (no aug.) &
  \multicolumn{5}{c|}{93.69} &
  \cellcolor[HTML]{EFEFEF}93.69 &
  \multicolumn{5}{c|}{94.64} &
  \cellcolor[HTML]{EFEFEF}94.64 \\ \hline
ReMix(append) &
  93.85 &
  94.02 &
  94.10 &
  94.03 &
  \multicolumn{1}{c|}{93.86} &
  \cellcolor[HTML]{EFEFEF}93.97 &
  94.66 &
  95.13 &
  95.31 &
  95.28 &
  \multicolumn{1}{c|}{94.87} &
  \cellcolor[HTML]{EFEFEF}95.05 \\
ReMix(replace) &
  93.85 &
  94.18 &
  94.18 &
  93.54 &
  \multicolumn{1}{c|}{93.38} &
  \cellcolor[HTML]{EFEFEF}93.83 &
  93.82 &
  94.01 &
  94.47 &
  94.36 &
  \multicolumn{1}{c|}{94.34} &
  \cellcolor[HTML]{EFEFEF}94.20 \\
ReMix(interpolate) &
  93.78 &
  94.10 &
  94.19 &
  93.96 &
  \multicolumn{1}{c|}{93.86} &
  \cellcolor[HTML]{EFEFEF}93.98 &
  95.25 &
  95.08 &
  95.23 &
  95.15 &
  \multicolumn{1}{c|}{95.04} &
  \cellcolor[HTML]{EFEFEF}94.05 \\
ReMix(covary) &
  93.85 &
  94.10 &
  94.26 &
  94.10 &
  \multicolumn{1}{c|}{93.94} &
  \cellcolor[HTML]{EFEFEF}94.05 &
  94.84 &
  94.83 &
  95.25 &
  94.99 &
  \multicolumn{1}{c|}{94.41} &
  \cellcolor[HTML]{EFEFEF}94.86 \\ \hline
$\mathbb{E}(p|\mathrm{aug})$ &
  \cellcolor[HTML]{EFEFEF}93.83 &
  \cellcolor[HTML]{EFEFEF}94.10 &
  \cellcolor[HTML]{EFEFEF}94.18 &
  \cellcolor[HTML]{EFEFEF}93.91 &
  \multicolumn{1}{c|}{\cellcolor[HTML]{EFEFEF}93.76} &
  \cellcolor[HTML]{EFEFEF}93.96 &
  \cellcolor[HTML]{EFEFEF}94.64 &
  \cellcolor[HTML]{EFEFEF}94.76 &
  \cellcolor[HTML]{EFEFEF}95.06 &
  \cellcolor[HTML]{EFEFEF}94.95 &
  \multicolumn{1}{c|}{\cellcolor[HTML]{EFEFEF}94.67} &
  \cellcolor[HTML]{EFEFEF}94.82 \\
  \hline
\end{tabular}%
}
\end{table}

\begin{figure}[h]
\centering
\includegraphics[width=\textwidth]{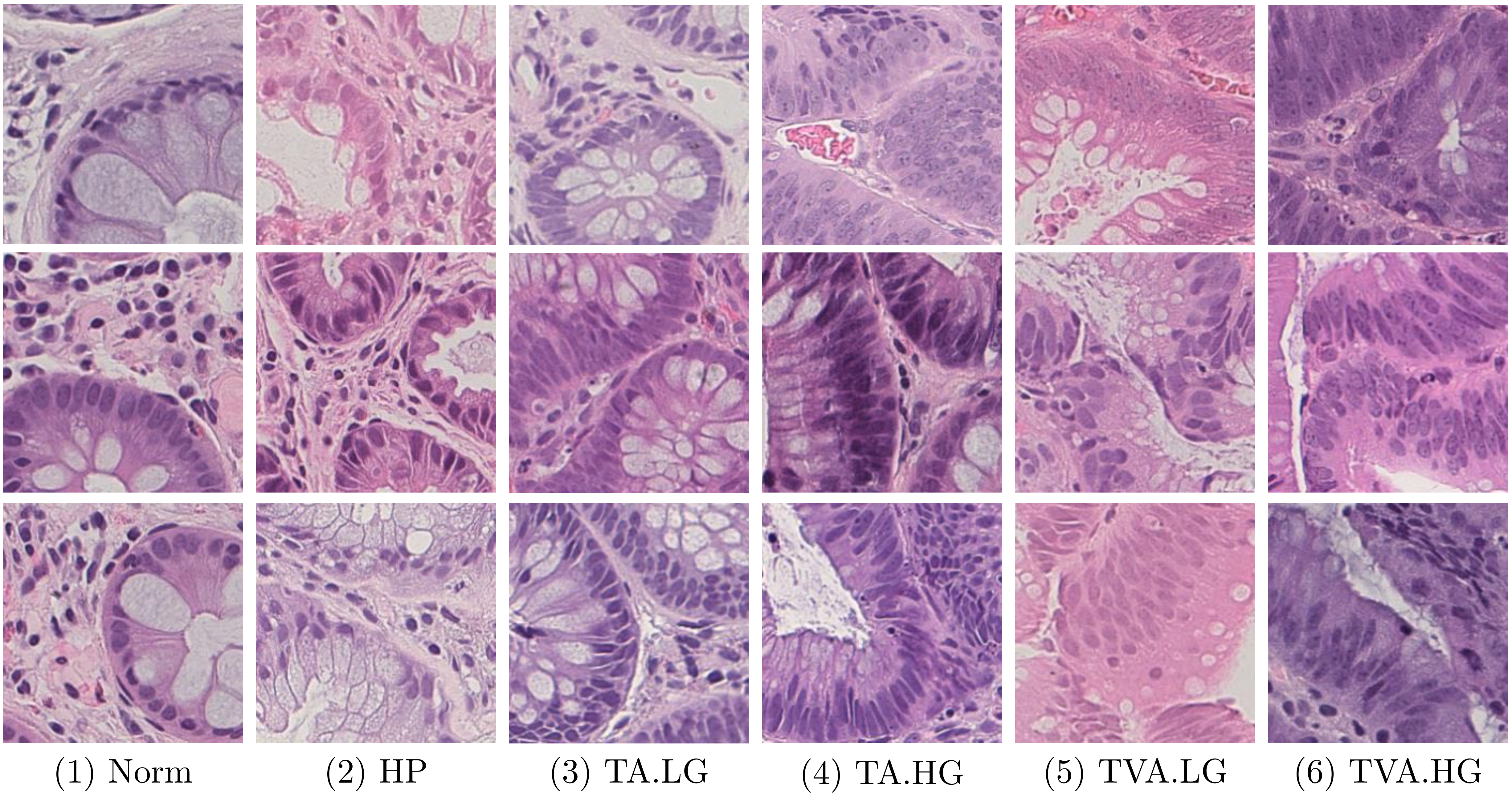}
\caption{\textbf{Representative patches in Unitopatho.} From randomly selected WSIs, we show the patches closest to their cluster centroids. Each patch is from a different WSI. Their difference indicates that even one prototype could convey enough information for certain whole-slide classification tasks. Six classes in this dataset are Normal tissue (NORM), Hyperplastic Polyp (HP), Tubular Adenoma with High-Grade dysplasia (TA.HG) and Low-Grade dyplasia (TA.LG), and Tubulo-Villous Adenoma with High-Grade dysplasia (TVA.HG), and Low-Grade dysplasia (TVA.LG).} 
\end{figure}

\end{document}